\documentclass[conference]{IEEEtran}
\IEEEoverridecommandlockouts
\usepackage{cite}
\usepackage{amsmath,amssymb,amsfonts}
\usepackage{booktabs} 
\usepackage{algorithmic}
\usepackage{graphicx}
\usepackage{textcomp}
\usepackage{xcolor}
\usepackage{hyperref}
\usepackage{amsmath}
\usepackage{graphicx}
\usepackage{algorithm}
\usepackage{algorithmic}
\usepackage{tabularx}
\usepackage{multirow,multicol}
\usepackage{makecell}
\usepackage{subcaption}
\usepackage{blindtext}
\usepackage{tabularx}
\usepackage{capt-of}
\usepackage{wrapfig}
\usepackage{enumitem}
\usepackage{xspace}
\def\BibTeX{{\rm B\kern-.05em{\sc i\kern-.025em b}\kern-.08em
    T\kern-.1667em\lower.7ex\hbox{E}\kern-.125emX}}
    \newcommand{\al}{\texttt{AutoLabel}~}
\newcommand{\aug}{\texttt{Aug}}
\begin{document}

\title{What Are Effective Labels for Augmented Data?  \\Improving Calibration and Robustness with AutoLabel\\
}

\author{\IEEEauthorblockN{Yao Qin}
\IEEEauthorblockA{\centering{\textit{Google Research}}}
\and
\IEEEauthorblockN{Xuezhi Wang}
\IEEEauthorblockA{\centering{\textit{Google Research}}}
\and
\IEEEauthorblockN{Balaji Lakshminarayanan}
\IEEEauthorblockA{\centering{\textit{Google Research}}}
\and
\IEEEauthorblockN{Ed H. Chi}
\IEEEauthorblockA{\centering{\textit{Google Research}}}
\and
\IEEEauthorblockN{Alex Beutel}

\IEEEauthorblockA{\centering{\textit{Google Research}}
}


}

\maketitle

\begin{abstract}
A wide breadth of research has devised data augmentation approaches that can improve both accuracy and generalization performance for neural networks. However, augmented data can end up being far from the clean training data and what is the appropriate label is less clear. Despite this, most existing work simply uses one-hot labels for augmented data. In this paper, we show re-using one-hot labels for highly distorted data might run the risk of adding noise and degrading accuracy and calibration. To mitigate this, we propose a \emph{generic} method \al to automatically learn the confidence in the labels for augmented data, based on the transformation distance between the clean distribution and augmented distribution. 
\al is built on label smoothing and is guided by the calibration-performance over a hold-out validation set.
We successfully apply \al to three different data augmentation techniques: the state-of-the-art RandAug, AugMix, and adversarial training.
Experiments on CIFAR-10, CIFAR-100 and ImageNet show that \al significantly improves existing data augmentation techniques over models' calibration and accuracy, especially under distributional shift. 
Additionally, \al improves adversarial training by bridging the gap between clean accuracy and adversarial robustness.
\end{abstract}


\section{Introduction}
Deep neural networks are increasingly being used in high-stakes applications such as healthcare and autonomous driving. For safe deployment, we not only want models to be accurate on independent and identically distributed (i.i.d.) test cases, but we also want models to be robust to distribution shift \cite{amodei2016concrete} and to not be vulnerable to adversarial attacks \cite{Goodfellow2014ExplainingAH, carlini2017towards, madry2017towards, qin2019detecting}. Recent work has shown that the accuracy of state-of-the-art models drops significantly when tested on corrupted data \cite{hendrycks2019benchmarking}.  Furthermore, these models do not just perform worse on these unexpected examples, but are also over-confident -- \cite{ovadia2019can} showed that calibration of models degrades under shift. Calibration measures the gap between a model's own estimate of correctness (i.e., confidence) versus the empirical accuracy, which measures the actual probability of correctness.  When a model is not well calibrated, particularly on unexpected examples, it undermines our ability to trust its predictions. 
Building models that are accurate \emph{and} robust, i.e. can be \textit{trusted} under unexpected inputs from both distributional shift and adversarial attacks, is a challenging but important research problem.

While numerous approaches have been explored for improving both calibration under distribution shift and adversarial robustness, one of the fundamental building blocks is \textit{data augmentation}: generating synthetic examples, typically by modifying existing training examples, that provide additional training data outside the empirical training distribution.
A wide breadth of literature has explored what are effective ways to modify training examples, such as making use of domain knowledge through label-preserving transformations \cite{augmix} or adding adversarially generated, imperceptible noise \cite{madry2017towards, trades}.
Approaches like these have been shown to improve the robustness and calibration of overparametrized neural networks as they alleviate the issue of neural networks overfitting to spurious features that do not generalize beyond the i.i.d. test set.  
\begin{figure*}[ht!]
     \centering

      \includegraphics[width=0.95\linewidth]{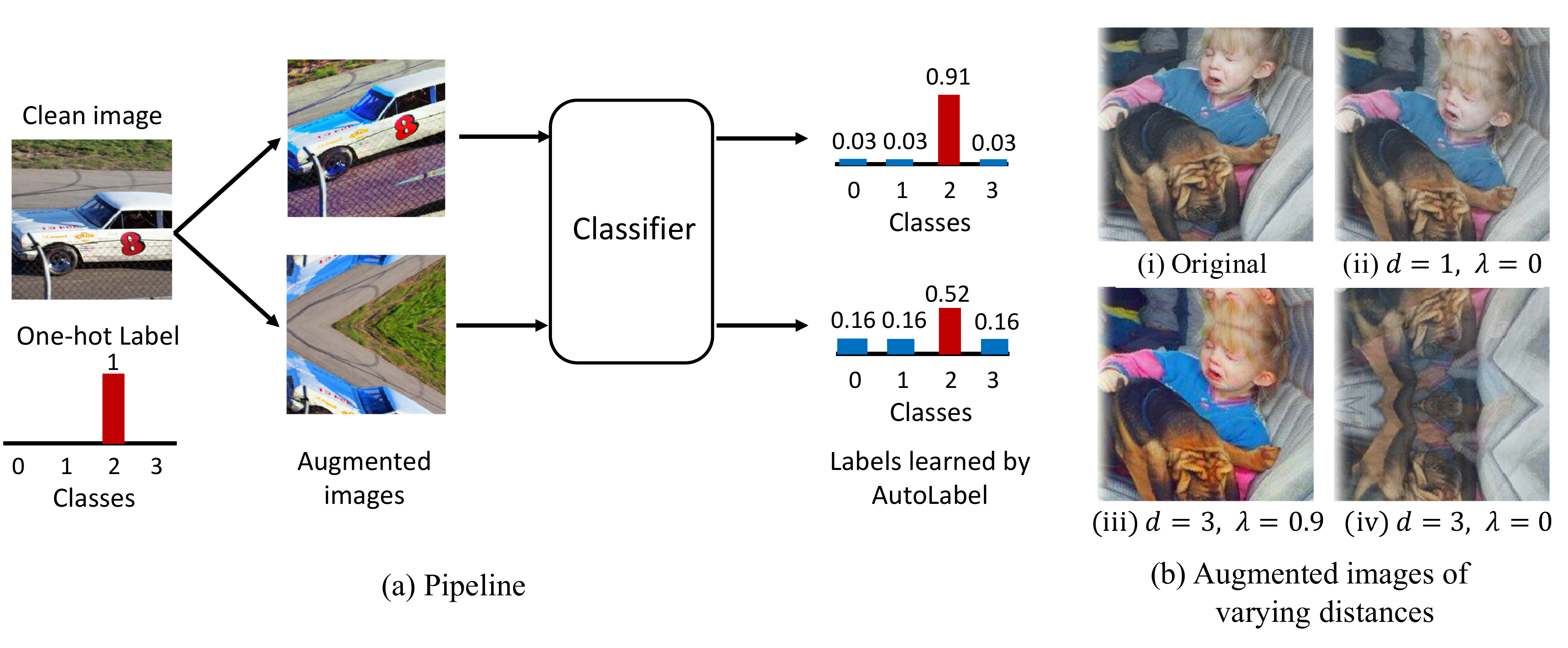}

     \caption{(a): An example showing \al assigning different labels to augmented images (e.g., by AugMix~\cite{augmix}) based on their transformation distances to the clean image. The label for the true class is automatically learned based on the calibration performance on validation set. (b): Examples of images augmented by AugMix with different distances to the original image.}
  \label{fig:pipeline:augmix}

\end{figure*}

In the broad amount of research on data augmentation, most of it attempts to apply transformations that do not change the true label such that the label of the original example can also be assumed to be the label of the transformed example, without expensive manual review.
While there has been a significant amount of work in how to construct such pseudo-examples in \textit{input} space, there has been relatively little attention on whether this assumption of label-preservation holds in practice and what \textit{label} should be assigned to such augmented inputs. For instance, many popular methods assign one-hot targets to both training data as well as augmented inputs that can be quite far away from the training data where even human raters may not be 100\% sure of the label. This runs the risk of adding noise to the training process and degrading accuracy and calibration, as the model may learn to assign high confidence predictions to inputs far away from training data. 

With this observation, in this paper we investigate the confidence assigned to target labels for augmented inputs and propose \al, a method that automatically adapts the confidence assigned to augmented labels, assigning high confidence to inputs close to the training data and lowering the confidence as we move farther away from the training data.  Figure~\ref{fig:pipeline:augmix} (left) gives a high-level overview of our proposed \al along with examples of augmented images of varying distances generating by AugMix~\cite{augmix} on the right.  

In summary, our key contributions are: 
\begin{itemize}
    \item We propose \al, a \emph{generic} approach that can automatically learn the confidence in labels for augmented data based on their transformation distance. 
    \item We show that \al is complementary to methods which focus on generating augmented inputs by combining it with RandAug~\cite{cubuk2020randaugment} (which includes 10 different augmentation types), the state-of-the-art method  AugMix~\cite{augmix}, as well as adversarial training~\cite{madry2017towards}. 
    \item We perform experiments on CIFAR-10, CIFAR-100 and ImageNet demonstrating that \al significantly improves the calibration of models 
    on both clean and corrupted data. In addition, \al improves adversarial training via bridging the gap between accuracy and adversarial robustness.
\end{itemize}

\vspace{2mm}
\section{Motivation: Why do we need AutoLabel?}

Before we present our method, we  investigate a key question:

\vspace{2mm}
 \emph{
How does the confidence assigned to labels of highly distorted augmented data affect model performance?
 } 
\vspace{2mm}
Using highly distorted augmented data is typically beneficial for improving robustness of the model and improves accuracy under covariate shift. However, using one-hot labels for highly distorted augmented data might decrease the clean accuracy as we assign 100\% confidence to both clean training data and highly distorted training data. 
We hypothesize that tuning the confidence assigned to distorted augmented data using \al\ avoids this trade-off and increases both clean accuracy and accuracy under distribution shift.

\begin{figure*}[t!]
     \centering
      \includegraphics[width=1\linewidth]{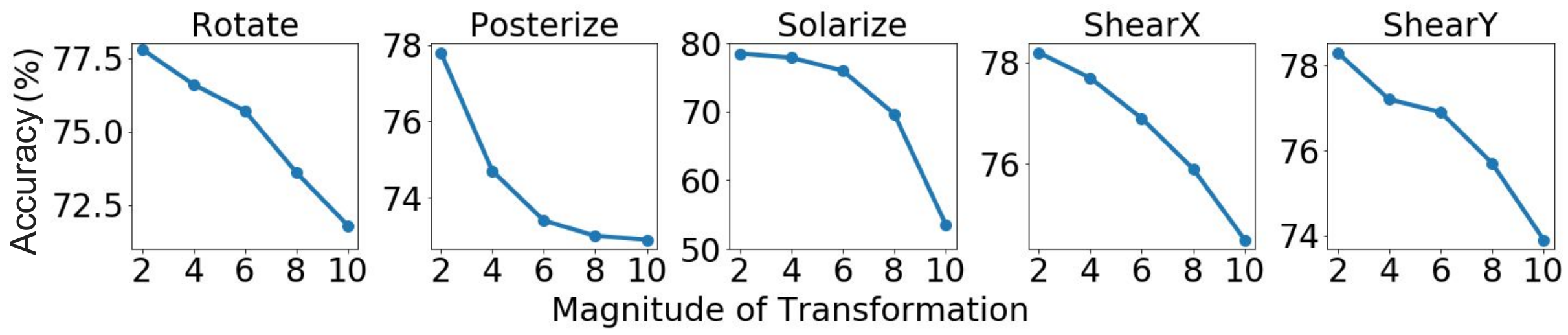}
     
     \caption{Test accuracy on CIFAR-100 test dataset to which each transformation of a specific magnitude is applied. Lower accuracy indicates higher distortion on the transformed images.}
  \label{fig:5augs}

\end{figure*}

To test this hypothesis,  we first train a Wide ResNet-28-10~\cite{Zagoruyko2016WideRN} on CIFAR-100~\cite{krizhevsky2009learning} 
and then report the accuracy on the test set with each transformation of different magnitudes applied. We use five different transformations: rotation, posterize, solarize, shear X and shear Y, whose distortion degree increases monotonically with the magnitude of the transformation, as used in \cite{cubuk2019autoaugment}.
We use the test accuracy of a transformation at a specific magnitude as an approximation of the distortion degree of the transformed images: lower test accuracy usually indicates higher degree of distortion on the transformed images.
 Figure~\ref{fig:5augs} shows that test accuracy monotonically drops with increasing distortion magnitude; when such highly distorted data are used during training with one-hot labels, we may add a lot of noisy inputs which may lead to overfitting and potentially hurt clean accuracy. 

\begin{figure*}[ht!]
     \centering
      \includegraphics[width=1\linewidth]{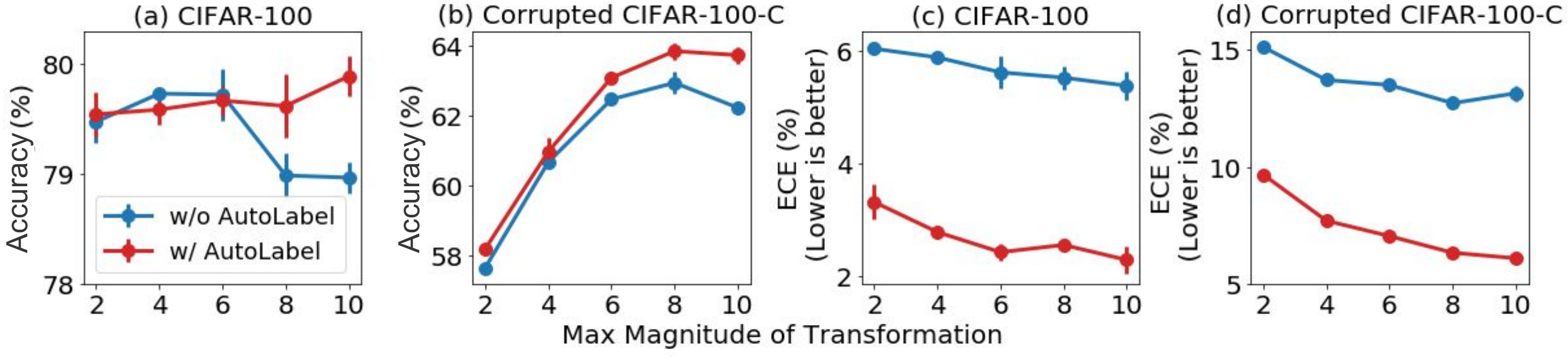}
     \caption{\textbf{Large distortions limit or hurt accuracy and calibration:} Test accuracy (higher is better) and expected calibration error (ECE, lower is better) on CIFAR-100 test dataset and CIFAR-100-C corruption dataset. We randomly sample a transformation from rotation, posterzie, solarize, shear X and shear Y with a magnitude ranging from 1 to the max magnitude and apply this transformation to the training image for data augmentation. As the $x$ axis increases, we include transformed images with higher distortion during training. The results are based on 5 independent runs.}
  \label{fig:acc_ece}
\end{figure*} 

Next, we train two networks using the above five transformations to augment the training data. The first network is trained with one-hot labels, another is trained with \al (which we will introduce in Section~\ref{sec:autolabel}), which automatically adjusts the labels for each magnitude of each transformation. All the other training details are kept the same for these two networks. During training, we randomly sample a transformation with a magnitude ranging from 1 to the max magnitude and then apply this transformation to the training image for data augmentation.
We report test accuracy and expected calibration error (ECE)~\cite{Guo2017OnCO},
which measures how well aligned the average accuracy and the average predicted confidence, on CIFAR-100 as well as CIFAR-100-C~\cite{hendrycks2019benchmarking}, which includes 17 different corruption types. {Note that there is \emph{no overlap} between the 5 transformations used for data augmentation and the 17 corruption types in CIFAR-100-C.}

In Figure~\ref{fig:acc_ece}, we report the accuracy (higher is better) and calibration (lower ECE is better) as the max magnitude of transformation increases (more highly distorted augmented data are incorporated into training).
For networks trained without \al i.e., using one-hot labels (shown in blue), as we can see from the first graph in Figure~\ref{fig:acc_ece}, the clean accuracy drops significantly when transformed images with higher distortion are added. On the other hand, the networks trained with \al (shown in red) can leverage highly distorted augmented data without incurring such a drop in clean accuracy.  
In addition, the model trained with \al consistently has a higher accuracy and a smaller calibration error on both CIFAR-100 and CIFAR-100-C (more significantly), and the gap widens as the max magnitude increases. This motivates the need for \al when using highly distorted augmented training data to improve model accuracy and calibration, especially under \textit{distributional shift}.

\section{Related Work}

\subsection{Data Augmentation}
Recent work has shown that introducing additional training examples can further improve a model's accuracy and generalization \cite{Devries2017ImprovedRO,cubuk2019autoaugment, Yun2019CutMixRS, Takahashi2019DataAU,  Lopes2019ImprovingRW,Zhong2020RandomED}. For example, AugMix \cite{augmix} utilizes stochasticity and diverse augmentations, together with a consistency loss over the augmentations, to achieve state-of-the-art corruption robustness.
Mixup \cite{Zhang2018mixupBE}, on the other hand, trains a neural network over convex combinations of pairs of examples and shows improved generalization of neural networks.
Furthermore, adversarial training \cite{Goodfellow2014ExplainingAH, madry2017towards, trades} can also be thought as a special data augmentation technique aiming for improving model's adversarial robustness. 
In this paper, we investigate the choice of the target labels for augmented inputs and show how to apply \al to these existing data augmentation techniques to further improve model's robustness.

\subsection{Calibration and Uncertainty Estimates}
A variety of methods have been developed for improving a model's calibration, e.g., post-hoc calibration by temperature scaling \cite{Guo2017OnCO} and multiclasss Dirichlet calibration \cite{class_calibration}.
Model's predictive uncertainty can also be quantified using Bayesian neural networks and approximate Bayesian approaches, e.g., variational inference \cite{NIPS2011_4329, pmlr-v37-blundell15}, MCMC sampling
based on stochastic gradients \cite{mcmc}, and dropout-based variational inference~\cite{kingma2015variational, gal2016dropout}. In addition to calibration over in-distribution data, more recently, \cite{ovadia2019can} show that model calibration can further degrade under unseen data shifts, where ensemble of deep neural networks \cite{ensemble} is shown to be most robust to dataset shift. On the other hand, several data augmentation methods have also been shown to improve model's calibration under data shifts. For example, AugMix is shown to improve uncertainty measures on corrupted image classification benchmarks \cite{augmix}. \cite{Thulasidasan2019OnMT} demonstrate that neural networks trained with mixup are significantly better calibrated under dataset shift, and are less prone to over-confident predictions on out-of-distribution data.

\subsection{Label Smoothing} 
Label smoothing, initially proposed in \cite{Szegedy2016RethinkingTI}, is used to prevent a model from being too over-confident in its predictions, thus improving its generalization ability.
It has been shown by \cite{Mller2019WhenDL,Thulasidasan2019OnMT} that label smoothing can also effectively improve the quality of a model's uncertainty estimates.
Our work is most closely related to the adaptive label smoothing algorithm in \cite{qin2020improving}.  \cite{qin2020improving} observe the connection between adversarial robustness and uncertainty, and propose an algorithm for adaptively updating the amount of label smoothing based on the adversarial vulnerability of \emph{clean data} to improve model's calibration.
In contrast, we propose to adaptively smooth the labels for \emph{augmented data} based on the distance to the clean training data, and show it can further improve a model's accuracy, calibration and adversarial robustness.
\vspace{4mm}
\section{AutoLabel: A Generic Framework for Setting Labels on Augmented Data}\label{sec:autolabel}
\subsection{Notation} Given a clean dataset $\mathcal{D} = \{(x_i, y_i)\}_{i=1, \cdots, m}$, where $y\in\{1,..., K\}$, the one-hot encoding of the label is denoted as $\hat{y} \in \{0, 1\}^K$, where the label for the true class $\hat{y}_{k=y} = 1$ and $\hat{y}_{k\neq y} = 0$ for others. In addition to the training data $\mathcal{D}$, we also have a clean validation set $\mathcal{D}_V$ drawn i.i.d. from the same distribution.

\subsection{The \al Algorithm}

Our key insight is that the \textbf{\emph{confidence}} in the labels associated with the augmented data likely depends on how distorted the transformation is. To make use of this insight, we would like to assign different confidence values in labels for the augmented training data based on their transformation distances. Many data augmentation approaches have hyperparameters that reflect how large the transformation should be. As examples, which we will discuss in-depth below, this can take the form of the number of transformations in AugMix~\cite{augmix} or the norm of the adversarial perturbation in \cite{madry2017towards} as shown in Table~\ref{tab:distance}. With aware of the transformation distance, then the problem becomes: 

\vspace{2mm}
\emph{How should we set the confidence value in the labels for augmented data? } 
\vspace{2mm}

\begin{figure}[t!]
     \centering
      \includegraphics[width=1.\linewidth]{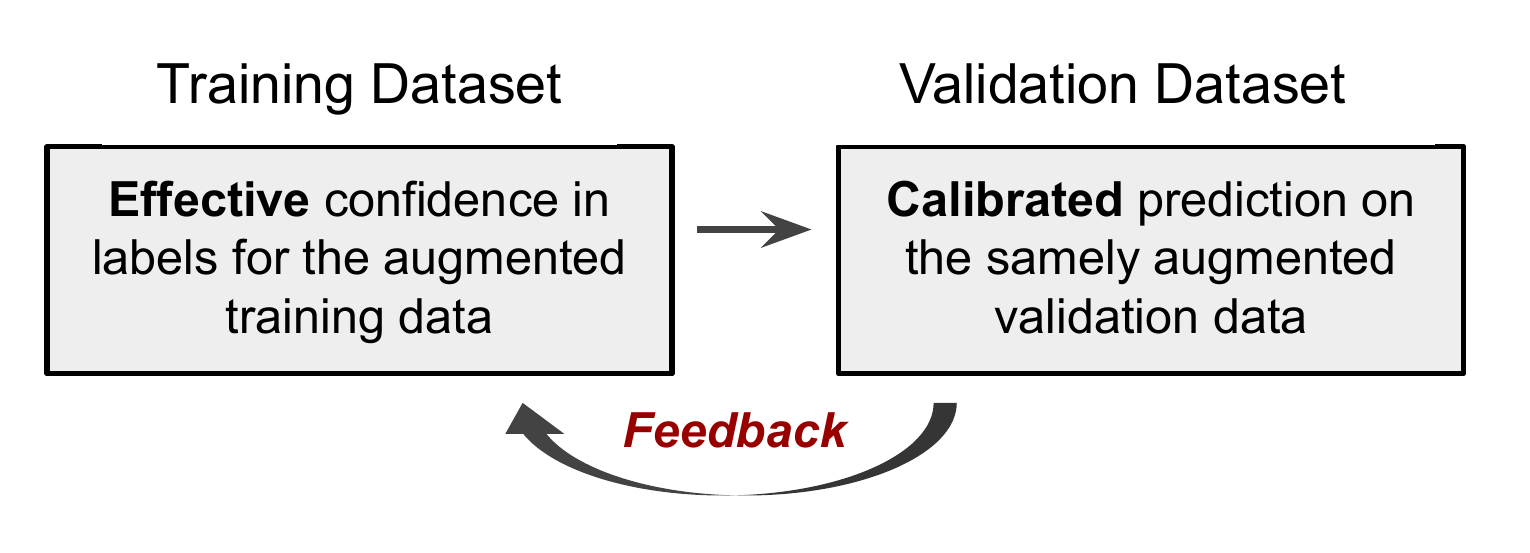}
     
     \caption{
A schematic diagram for updating confidence of training labels in \al algorithm.}
  \label{fig:alg}

\end{figure}

To address this problem, we build \al upon the hypothesis that {effective labels for {augmented training data} can lead to well-calibrated predictions on the \textbf{samely} augmented validation set}. Thus, the training labels for augmented data can be automatically updated according to the calibration performance on the samely augmented validation data. A schematic diagram for updating confidence of training labels in \al is shown in Figure~\ref{fig:alg}. Specifically, if a model is over-confident on the augmented validation set, then the confidence in the training labels should be decreased accordingly; otherwise the confidence should be increased. Note that computing calibration on the augmented validation set does not use the confidence in the labels of validation data.
\begin{table*}[t]
\caption{Transformation distance used by \al for each data augmentation method.}

    \centering
    \begin{tabularx}{0.9\textwidth}{c|X}
        \toprule \midrule
         \textbf{Augmentation Method} & \makecell{\textbf{Transformation distance determined by}} \\
         \midrule
         RandAug & \makecell{transformation type,  magnitude of transformation $m$}  \\
         \midrule
         AugMix & \makecell{depth of augmentation chain $d$, mixing parameter $\lambda$}  \\
         \midrule
         Adversarial Training & \makecell{ max $\ell_\infty$ norm $\epsilon$}\\
         \midrule
         \bottomrule
    \end{tabularx}
    \label{tab:distance}
\end{table*}

Taking these together, our proposed \al is mainly composed of two components: 
\begin{enumerate}
    \item a measure of the transformation distance for the augmented data,
    \item a subroutine for updating labels of the augmented data during training.
\end{enumerate}

Specifically, given a data augmentation technique $\aug$ that takes in an image $x$ and outputs an augmented image $\aug(x, s)$ that transformed by a distance $s \in \mathrm{R}$, \al updates its label based on 
the calibration performance on the samely augmented validation data. Since calibration can not be computed over a single data point, we must obtain an augmented validation set that is transformed by the same distance $s$.
To this end, we discretize the transformation distance $s$ into $N$ buckets $\{S_1, \cdots, S_N\}$ where each $S_n$ is a range, and we can generate augmented data for bucket $S_n$ by sampling a distance uniformly in that range $s \sim \mathcal{U}(S_n)$ to generate $\aug(x,s)$. In this way, we can generate the augmented validation set $\mathcal{Q}(S_n) = \{(\aug(x_i, s), y_i) | (x_i, y_i) \in \mathcal{D}_V, s \sim \mathcal{U}(S_n)\}$,
which is used to learn the labels for any training data transformed by a distance $s \in S_n$.

With the augmented validation set $\mathcal{Q}(S_n)$, \al updates the confidence of the true class $\Tilde{y}_{k=y}(S_n)$ after each training epoch $t$ according to:
\vspace{-1mm}
\begin{multline}\label{eqn:update}
    \Tilde{y}_{k=y}^{t+1}(S_n) =  \Tilde{y}_{k=y}^{t}(S_n) - \alpha \cdot {\text{ECE}}^t(\mathcal{Q}(S_n))\\  \cdot  \text{sign}(\text{Conf}^t(\mathcal{Q}(S_n)) - \text{Acc}^t(\mathcal{Q}(S_n)))
\end{multline}
where ECE$(\mathcal{Q}(S_n))$, $\text{Acc}(Q(S_n))$ and $\text{Conf}(\mathcal{Q}(S_n))$ are respectively the expected calibration error, accuracy and confidence on the augmented validation set.
The sign of $($Conf$(\mathcal{Q})$ - Acc$(\mathcal{Q}))$ indicates if the model is overall over-confident ($>0$) or under-confident ($<0$). Intuitively, if the model is over-confident on the validation set, we should reduce the confidence given to the true class $\Tilde{y}_{k=y}$, otherwise we should increase $\Tilde{y}_{k=y}$. The expected calibration error on the augmented validation set $\text{ECE}(\mathcal{Q}) \geq 0$ suggests to what extent we should adjust the labels as the optimal result is $\text{ECE}(\mathcal{Q}) = 0$ when the training converges. The hyperparameter $\alpha$ controls the step size of updating the labels. Since $\Tilde{y}^{t+1}_{k=y}$ stands for the probability of the true class, we clip the value to be within $[\text{Acc}^t(\mathcal{Q}), 1]$ after each update. $\text{Acc}^t(\mathcal{Q})$ is used as the minimum clipping value to prevent $\Tilde{y}^{t+1}_{k=y}$ from being too small as $\text{Acc}^t(\mathcal{Q}) \rightarrow \frac{1}{K}$ when the classifier is a random guesser.

Given the updated label for the true class $\Tilde{y}^{t+1}_{k=y}(S_n)$, \al takes a label smoothing approach to uniformly distribute the remaining probability to other classes:
\begin{equation}\label{eqn:other}
    \Tilde{y}_{k\neq y}^{t+1}(S_n) = (1 - \Tilde{y}_{k= y}^{t+1}(S_n)) \cdot \frac{1}{K - 1},
\end{equation}
where $K$ is the number of classes in the dataset and $\mathop{\sum}_{k=1}^K\Tilde{y}_k = 1$. 
Finally, \al trains the model using $\Tilde{y}(S_n)$ as the target for the cross-entropy loss across the augmented data. A complete pseudocode for \al is presented in Algorithm 1. 
\begin{algorithm*}[ht]
  \caption{Pseudocode of \al}
  \label{al_algo}
\begin{algorithmic}[1]
\STATE \textbf{Input:} A training dataset $\mathcal{D} = \{(x_i, y_i)\}_{i=1, \cdots, m}$, a validation dataset $\mathcal{D_V}$ drawn i.i.d. from the same distribution, an augmentation method $\aug$.
Number of classes $K$, number of training epochs $T$, number of distance buckets $N$ and the hyperparameter $\alpha$. 
\STATE We perform \aug~ to obtain the augmented training data $\aug(x, s)$, where the transformation distance $s$ is determined by the hyperparameters in the \aug. We discretize the transformation distance $s$ into $N$ buckets $\{S_1, \cdots, S_N\}$, where each $S_n$ is a range.
\STATE 
For each distance bucket $S_n$, we initialize  $\Tilde{y}^0(S_n)$ as the one-hot label.
  \FOR{$t=0$ {\bfseries to} $T-1$}
  \STATE Minimize cross-entropy loss over the augmented training data with smoothed labels $\Tilde{y}^t(S_n)$.
  \FOR{$n=1$ {\bfseries to} $N$}
  \STATE Generate an augmented validation set: $\mathcal{Q}(S_n) = \{(\aug(x_i, s), y_i)|(x_i, y_i) \in \mathcal{D}_V, s\sim \mathcal{U}(S_n)\}$.
  \STATE Update the label for the true class $\Tilde{y}_{k=y}^{t+1}(S_n)$:  
   
  $\Tilde{y}_{k=y}^{t+1}(S_n) =  \Tilde{y}_{k=y}^{t}(S_n) - \alpha \cdot {\text{ECE}}^t(\mathcal{Q}(S_n)) \cdot \text{sign}(\text{Conf}^t(\mathcal{Q}(S_n)) - \text{Acc}^t(\mathcal{Q}(S_n)))$ \hfill $\triangleright$ \textit{according to Eqn (1)}
  \STATE Clip $\Tilde{y}^{t+1}_{k=y}(S_n)$ to be within $[\text{Acc}^t(\mathcal{Q}(S_n)), 1]$
  \STATE Update the label for other classes $\Tilde{y}_{k\neq y}^{t+1}(S_n)$:
  $\Tilde{y}_{k\neq y}^{t+1}(S_n)= (1 - \Tilde{y}_{k= y}^{t+1}(S_n)) \cdot \frac{1}{K - 1}$
  \hfill $\triangleright$ \textit{according to Eqn (2)}
  \ENDFOR
  \ENDFOR
\end{algorithmic}
\end{algorithm*}

\vspace{2mm}
\section{AutoLabel + Data Augmentations}
\vspace{2mm}
To demonstrate \al can easily slot into existing data augmentation methods, we show how to apply \al to automatically adjust the confidence in labels over different data augmentation methods: 
\begin{itemize}
    \item RandAug~\cite{cubuk2020randaugment}: including 10 different types of simple transformations used by AutoAugment~\cite{cubuk2019autoaugment}. Note that these transformations \textbf{do not overlap} with corruption types in the test corrupted datasets.
    \item AugMix~\cite{augmix}: Mixing diverse simple augmentations in convex combinations.
    \item Adversarial Training~\cite{madry2017towards}: A special case of data augmentation to improve adversarial robustness via training on constructed adversarial examples.
\end{itemize}

These data augmentation methods originally use one-hot labels for augmented data and we discuss in details how to apply \al for each of them.\footnote{We also apply \al to mixup~\cite{Zhang2018mixupBE}, which is a data augmentation technique that uses soft-labels, and observe an improvement of \al over calibration. We refer interested readers to our supplementary material for more details and results.} 
Table~\ref{tab:distance} shows an overview of how \al differentiates the augmented data based on the transformation distance under each data augmentation method.

Below we first give an overview of each data augmentation method and introduce how to differentiate the augmented data based on their distance to the clean distribution, then we show how to apply \al to learn more appropriate labels.
\subsection{AutoLabel for RandAug}

RandAug includes 10 different types of transformations from AutoAugment~\cite{cubuk2019autoaugment} and RandAugment~\cite{cubuk2020randaugment}: color, rotation, autocontrast, equalize, posterize, solarize, shear X, shear Y, translate X and translate Y. 
During training, RandAug randomly applies one transformation to generate the augmented data. Unlike RandAugment~\cite{cubuk2020randaugment} that optimizes a single distortion magnitude for all the transformations, RandAug randomly samples a distortion magnitude $m \in \{1, \cdots, m_{max}\}$, where $m_{max}$ is the maximum distortion magnitude. Finally, the model is trained on the augmented data with one-hot labels.

Instead of using one-hot labels, we propose \al to automatically learn the confidence in labels for the augmented data that are transformed by different transformation distances. 
In RandAug, 
the transformation distance is determined by two factors: (1) the type of sampled transformation, in total we have 10 different transformations, and (2) the distortion magnitude $m$. 

To learn the labels for the augmented data transformed by a specific operation with a distortion magnitude at $m$, \al applies the same transformation distorted by the same magnitude $m$ to construct the augmented validation set. Then \al updates the confidence of labels according to Eqn~(\ref{eqn:update}) $\&$ (\ref{eqn:other}) and trains the model with these updated labels.

\subsection{AutoLabel for AugMix}
AugMix~\cite{augmix}
is a data augmentation technique that achieves state-of-the-art robustness and uncertainty estimates under data shift. Specifically, AugMix 
augments the input data via feeding the input $x$ into an augmentation chain\footnote{The original AugMix~\cite{augmix} uses 3 augmentation chains. However, we consistently observe an accuracy increase when we use one augmentation chain. 
} 
which consists of $d \in \{1, 2, 3\}$ transformations randomly sampled from 10 different operations used in RandAug with a fixed distortion magnitude. Then a convex combination is performed to mix the augmented image $x_{\text{aug}}$ with the original image $x$: $\aug_{augmix}(x) = \lambda \cdot x + (1 - \lambda) \cdot x_{\text{aug}}$,
where the mixing parameter $\lambda \in [0,1]$ is randomly sampled from a uniform distribution.

In AugMix, the transformation distance is mainly controlled by two parameters\footnote{Transformation types could provide us more precise transformation distance but is not our main focus in AugMix. 
}: (1) the depth of the augmentation chain $d$, which decides how many augmentation operations are applied to the original image; (2) the mixing parameter $\lambda$, which controls the ratio of the augmented image $x_{\text{aug}}$ and the original image $x$.
In Figure~\ref{fig:pipeline:augmix}(b) (ii) and (iv), 
a deeper augmentation chain causes the image to quickly degrade and drift off the data manifold. In addition, when comparing the augmented images with different mixing parameter $\lambda$, shown in Figure~\ref{fig:pipeline:augmix}(b) (iii) and (iv), we can see that as $\lambda \rightarrow 0$, the augmented image is further away from the clean image. 
 As a result, we can define the distance bucket $S_{d,n}$ for the augmented data as:
    $S_{d,n} = S_{d, \lceil \lambda N\rceil},$\footnote{In the special case where $\lambda = 0$, we merge it into bucket $S_1$ to avoid creating an additional bucket, similarly for $\epsilon$ in adversarial training.} where $N$ is the total number of buckets at a given depth $d$.

Next, to learn the labels for augmented training data within a distance bucket $S_{d,n}$, \al constructs an augmented validation set $\mathcal{Q}(S_{d,n})$ by feeding the validation images into an augmentation chain with the depth $d$ and then randomly sample a mixing parameter $\lambda'$ from a uniformly distribution: $\lambda' \sim \mathcal{U} (\frac{n}{N}, \frac{n+1}{N})$ to mix the original image and the augmented image. Finally, \al updates the labels $\Tilde{y}(S_{d,n})$ according to Eqn~(\ref{eqn:update}) $\&$ (\ref{eqn:other}) and trains the model using these updated labels.

\subsection{AutoLabel for Adversarial Training}\label{sec:adv}
Adversarial training~\cite{Goodfellow2014ExplainingAH} can be formulated as solving the min-max problem: 
\begin{equation}\label{eqn:adv}
    \mathop{\min}_w \mathop{\mathbb{E}}_{||\delta||_\infty \leq \epsilon}[\max \mathcal{L}(f(x+\delta;w), y)],
\end{equation}
where $\delta$ denotes the adversarial perturbation, $\epsilon$ denotes the maximum $\ell_\infty$ norm of adversarial perturbation and a one-hot encoding of the label $y$ is used as the target for the cross-entropy loss $\mathcal{L}$. In \cite{madry2017towards}, the inner maximization problem is approximately solved by generating projected gradient descent (PGD) attacks. Therefore, standard adversarial training can be considered as a specific data augmentation that aims for improving model's adversarial robustness.

Instead of training a model with one-hot labels, \al differentiates the adversarial examples according to the distance between the adversarial examples and clean data, which is approximately captured by the
$\ell_\infty$ norm of the adversarial perturbation $\epsilon$. Unlike \cite{madry2017towards} using a fixed $\epsilon$ to construct PGD adversarial attacks, we randomly sample $\epsilon$ from a uniform distribution $\epsilon \sim \mathcal{U}(0, \epsilon_{max})$ to construct PGD adversarial attacks with different distances to the original data.
If the $\ell_\infty$ norm of the adversarial perturbation is bounded by $\epsilon$, then the constructed adversarial example falls into the distance bucket $S_n = S_{\lceil \epsilon \cdot \frac{N}{\epsilon_{max}}\rceil}$, where $N$ is the total number of distance buckets.

In order to learn the labels for adversarial examples within a distance bucket $S_n$, \al constructs adversarial examples for the validation images with the $\ell_\infty$ norm of the adversarial perturbation bounded by $\epsilon'$,
where $\epsilon'$ is randomly sampled from a uniform distribution:
  $ \epsilon' \sim \mathcal{U}( \frac{n \cdot \epsilon_{max}}{N}, \frac{(n+1)\cdot \epsilon_{max}}{N})$.
Finally the training labels for adversarial examples are updated following Eqn (\ref{eqn:update}) $\&$ (\ref{eqn:other}).

\vspace{3mm}
\section{
AutoLabel improves calibration
}
\vspace{1mm}
In this section, we mainly show how \al can help improve standard data augmentation techniques in terms of the calibration performance by assigning effective confidence in labels.
\subsection{Baselines} 
\subsubsection{{\textrm{Baselines for RandAug and AugMix}}} 
In order to demonstrate the effectiveness of \al while applying to existing data augmentation techniques, we use the state-of-the-art RandAug~\cite{cubuk2020randaugment} and AugMix~\cite{augmix} trained with one-hot labels as baselines. 

Further, as \al is built upon label smoothing (LS)~\cite{Szegedy2016RethinkingTI}, we also report the performance when label smoothing is applied to each of these data augmentations. 
\vspace{2mm}
\subsubsection{Baselines for adversarial training}\label{sec:baseline_at} In order to show \al can help adversarial training benefit calibration under distributional shift, we compare it with other 5 different models. They are
\begin{itemize}

    \item A vanilla model trained with one-hot labels. 
    \item Adversarial training (AT)~\cite{madry2017towards}, which is trained on projected gradient descent (PGD) attacks with one-hot labels. 
    \item Adversarial training with label smoothing (AT + LS). 
    \item Adversarial training with PGD attacks generated with multiple $\ell_\infty$ norm bounds ( AT + multiple $\epsilon$). Unlike standard adversarial training in~\cite{madry2017towards} using a fixed $\epsilon$ to construct PGD attacks, this model randomly samples an $\epsilon \in (0, \epsilon_{max}]$ during training. The one-hot labels are used for the generated adversarial attacks.
    \item Confidence-calibrated adversarial training (CCAT)~\cite{ccat}, which smooths the labels for adversarial examples according to $\Tilde{y} = g(\delta) \hat{y} + (1 - g(\delta) \frac{1}{K})$, where the balancing parameter $g(\delta)$ follows a  ``power transition'': $g(\delta):= (1 - \min(1, \frac{||\delta||_\infty}{\epsilon}))^\rho$, $\rho$ is set to 10 as \cite{ccat} to ensure a uniform distribution is used as the labels for adversarial examples when $||\delta||_\infty \geq \epsilon$. Similarly, $\epsilon$ is randomly sampled from (0, $\epsilon_{max}]$ during training. 
\end{itemize}

\subsection{Datasets} For RandAug and AugMix related experiments, we report the performance on 
CIFAR-100~\cite{krizhevsky2009learning} and ImageNet~\cite{Russakovsky2015ImageNetLS}. As adversarial training is hard to scale on ImageNet, we mainly perform experiments for adversarial training on CIFAR-10 and CIFAR-100~\cite{krizhevsky2009learning}. We randomly sample 5000 images from 50000 training data to serve as the hold-out validation set for both CIFAR datasets. For ImageNet, we evenly split 50000 test images into a validation set with 25000 images and test the models' performance on the hold-out test set with 25000 images.

In addition, we test models' robustness on the corrupted datasets: CIFAR-10-C, CIFAR-100-C, ImageNet-C~\cite{hendrycks2019benchmarking}, which include different corruptions types (17 types for CIFAR-10/100 and 15 types for ImageNet) that are frequently encountered in natural images. Each type of corruption has five corruption severities. Note that the corruption type in the corrupted dataset \textit{\textbf{do not overlap}} with transformations used for data augmentations.

\subsection{Evaluation Metrics} We report the classification accuracy and expected calibration error on the clean datasets as \textbf{Accuracy} (higher is better) and \textbf{ECE} (lower is better) respectively. Specifically, given a classifier $f(\cdot)$ for a $K$-class classification problem, let $f_k(x)$ denote the predicted probability for the $k$-th class. We use $f(x):= \text{argmax}_k f_k(x)$ to represent the predicted class and $c(x):= \max_k f_k(x)$ as model's confidence of the predicted class. The expected calibration error (ECE)~\cite{Guo2017OnCO, ovadia2019can} is defined as
$$\mathrm{ECE} = \sum_r^R \frac{|B_r|}{m} |\mathrm{acc}(B_r) - \mathrm{conf}(B_r)|,$$ 
where the input data is divided into $R$ buckets, $B_r$ indexes the $r$-th confidence bucket and $m$ denotes the data size, the accuracy and the confidence of $B_r$ are defined as:
$$\mathrm{acc}(B_r) = \frac{1}{|B_r|}\sum_{i\in B_r} {\bf{1}}(f(x_i) = y_i)$$ 
$$\mathrm{conf}(B_r) = \frac{1}{|B_r|}\sum_{i\in B_r} c(x_i)$$
Expected calibration error measures how well aligned the average accuracy and the average predicted confidence are. 

In addition, we also report the accuracy and expected calibration error on the corrupted datasets, denoted as \textbf{cAccuracy} and \textbf{cECE}, which are computed as an average over all the corruption types across 5 corruption severities. 
\subsection{Network Architectures and Hyperparameters}\label{sec:arch} 
We use a Wide ResNet-28-10~\cite{Zagoruyko2016WideRN} for both CIFAR-10 and CIFAR-100 datasets, and a ResNet-50~\cite{resnetv1} for ImageNet as our basic model architectures. We use the open-sourced code at \url{https://github.com/google/uncertainty-baselines/tree/master/baselines} to train all the models with the same training hyperparameters for fair comparison.  

Below we display all other hyperparameters involved for each method as well as \al.
\subsubsection{Label smoothing}
When applying label smoothing to RandAug, AugMix and adversarial training, we sweep the hyperparameter $\rho$ which decides the smoothing degree in a range [0, 0.1] with a step size 0.01 and find the best $\rho = 0.02$ for CIFAR10 and CIFAR100 and $\rho = 0.01$ for ImageNet. 
\subsubsection{Adversarial training} Similar to recent work \cite{xie2020adversarial}, we observe that training models with adversarial examples bounded with a smaller $\ell_\infty$ norm, e.g., $\epsilon_{max} = 0.01$, can benefit more to the corrupted accuracy with a small accuracy drop on the clean data. Therefore, we train all models with PGD attacks bounded by $\epsilon_{max} = 0.01$ updated with 10 iterations. The step size is set to be $\epsilon/4$.

\subsubsection{\al for RandAug} When \al is applied to RandAug, the hyperparameter $\alpha$ in Eqn (1) is swept in  [0, 0.1]. We choose the best $\alpha = 0.01$ for CIFAR-100 and $\alpha = 0.02$ for ImageNet based on the holdout validation set. The number of distance bucket is set to be $N=10$. 

\subsubsection{\al for AugMix}
Following original AugMix~\cite{augmix}, the max depth of the augmentation chain is set to be $d_{max}=3$ and the number of distance bucket is set to be $d_{max} \cdot N = 3 \cdot 5 = 15$.  We use the best $\alpha=0.02$ in Eqn (1) for CIFAR-100 and ImageNet.

\subsubsection{\al for adversarial training}
The number of distance buckets is set to be $N=10$ and the hyperparameter $\alpha$ in Eqn (1) is set to be $\alpha=0.5$ for CIFAR-10 and $\alpha=0.005$ for CIFAR-100.

\begin{table*}[t!]
    \caption{\textbf{Improvements of \al over RandAug and AugMix.} The accuracy and ECE are reported on both in-distribution test datasets (CIFAR-100 and ImageNet) as well as the corresponding corrupted datasets (CIFAR-100-C and ImageNet-C). All numbers reported in the table are in $\%$, an average of 4 independent runs on CIFAR-100 and 2 independent runs on ImageNet. The arrow indicates better direction. Best results are highlighted in \textbf{Bold}.}\label{tab: aug_result}
\centering
\begin{tabular*}{0.95\textwidth}{c @{\extracolsep{\fill}} cccc}
\toprule\midrule
\multirow{2}{*}{\textbf{Method}}                                             & \multicolumn{2}{c}{\textbf{Accuracy / cAccuracy} ($\uparrow$)}        & \multicolumn{2}{c}{\textbf{ECE / cECE} ($\downarrow$)}             \\ \cmidrule(r){2-3}  \cmidrule(r){4-5} 
                                                               & CIFAR-100   & ImageNet     & CIFAR-100-C   & ImageNet-C   \\
\midrule                                                              
{RandAug\cite{cubuk2020randaugment}}                                                           & 82.0 / 63.0    & 76.9 / 43.4 & 4.1 / 13.0         &  2.0 / 6.4 \\
{+ Label Smoothing}                                                           & 82.2 / 63.6    & 76.9 / 43.5& 2.4 / 8.2      & 1.2 / 5.5\\
{ + \al} & \textbf{82.7 / 64.8}       & \textbf{76.9 / 43.9}           & \textbf{1.9 / 5.6}        & \textbf{1.0 / 5.1}      \\ \midrule
{AugMix\cite{augmix}}                                                           & 81.1 / 64.3       & 75.9 / 46.1          & 4.5 / 10.9      & 1.5 / 4.9     \\
{+ Label Smoothing}                                                           & 81.3 / 64.6          &   76.0 / 46.3     & 2.6 / 6.5      &  1.5 / 4.6   \\
{ + \al} & \textbf{81.9 / 65.5}       & \textbf{76.4 / 46.5}           & \textbf{1.8 / 4.2}        & \textbf{1.4 / 4.2}      \\ 
\midrule
\bottomrule
\end{tabular*}
\end{table*}
\begin{figure}[t]
  \begin{center}
    \includegraphics[width=0.96\linewidth]{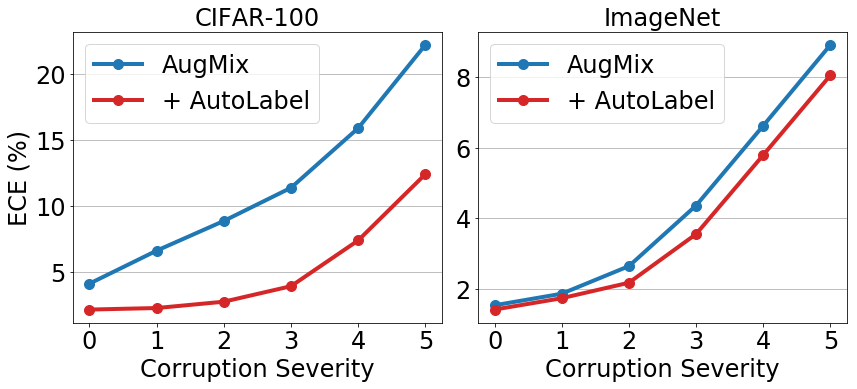}
  \end{center}
  \caption{Expected calibration error (ECE) of AugMix trained with one-hot labels and \al across corruption severities on CIFAR100 and ImageNet. Severity 0 denotes clean data. As the corruption severity increases, the improvement of \al increases.
}\label{fig:aug_cal}
\end{figure}

\subsection{Improvement over Calibration}
\subsubsection{\al improves calibration of RandAug and AugMix} In this section, we apply \al to RandAug and AugMix to investigate 
if \al can make data augmentation approaches more effective in improving calibration, especially under distributional shifts.  We see in Table~\ref{tab: aug_result} a clear picture: \al consistently helps RandAug and AugMix improve both accuracy and calibration across CIFAR100 and ImageNet, and greatly outperforms label smoothing. In addition, we can see that \al has a much more significant improvement on models' calibration on the corrupted datasets.  As shown in Figure \ref{fig:aug_cal} we analyze how calibration performance changes with the severity of the corruption being tested against, comparing AugMix trained with one-hot labels and with \al. We see that the baseline AugMix is increasingly worse calibrated as the corruption increases, but \al dampens that trend effectively. Similar patterns are observed when \al is applied to RandAug.

\begin{figure}[t!]

     \centering
      \includegraphics[width=0.96\linewidth]{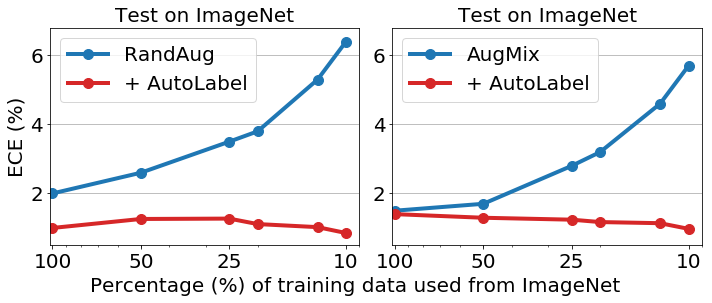}\\\vspace{3mm}
 \includegraphics[width=0.96\linewidth]{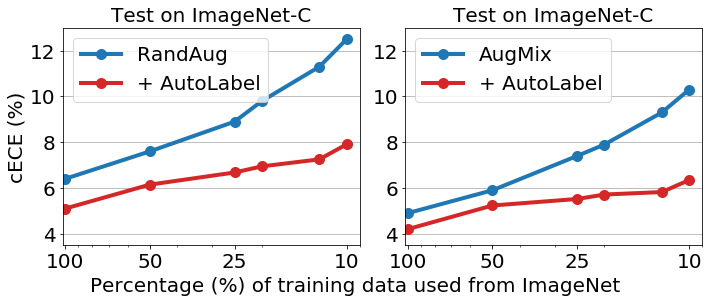}
      
     \caption{\textbf{\al improves calibration on ImageNet with a reduced size of training data.} When training a model with limited training data, the improvement of \al over calibration is significantly increased. $\alpha = 0.02$ is used for \al for all experiments here without further tuning.}
  \label{fig:size_imagenet}

\end{figure} 
Looking more closely, we find that the improvement of \al over calibration on ImageNet is relatively smaller compared to CIFAR-100. We conjecture that this is mainly due to the much larger training data on ImageNet, leading to a greater generalization performance and the headroom for improvement is relatively limited. To validate this, we train the same networks with a reduced size of training images, e.g., we randomly sample 50$\%$ training images from the whole training dataset ($\sim$1.2M training images). Note that we use the same hyperparameter $\alpha = 0.02$ in Eqn~(\ref{eqn:update}) without further tuning for each training data size. From Figure~\ref{fig:size_imagenet} we can see that as the percentage of training data is reduced, the calibration performance of both RandAug and AugMix trained with one-hot labels becomes significantly worse. In contrast, \al enables models to keep a low calibration error on the clean test set and a much smaller calibration error on the corrupted dataset. This indicates that \al could be especially beneficial to models' calibration performance when we have limited training data.

\begin{table*}[t]
\centering
\caption{\textbf{\al improves calibration over adversarial training.} Expected calibration error (ECE) on the clean CIFAR-10 and CIFAR-100 and on the corrupted CIFAR-10-C and CIFAR-100-C (measured by cECE). Note adversarial examples during training are bounded with a small $\ell_\infty$ norm ($\epsilon_{max} = 0.01$) for better in-distribution performance as~\cite{xie2020adversarial}. All the numbers reported are an average over 2 independents runs and in $\%$. Best result is highlighted in
\textbf{bold}.}\label{tab:at}
\begin{tabular*}{0.9\textwidth}{c @{\extracolsep{\fill}} cccc}
\toprule\midrule
\multirow{2}{*}{\textbf{Method}}                                             & \multicolumn{2}{c}{\textbf{Accuracy / cAccuracy} ($\uparrow$)}        & \multicolumn{2}{c}{\textbf{ECE / cECE} ($\downarrow$)}             \\ \cmidrule(r){2-3}  \cmidrule(r){4-5} 
                                                               & CIFAR-10   & CIFAR-100     & CIFAR-10-C   & CIFAR-100-C   \\
\midrule                                      

Vanilla & 95.6 / 76.0         & 79.5 / 52.0   & 2.6 / 15.8         & 6.1 / 17.6             \\ \midrule
{AT~\cite{madry2017towards}}  & 93.6 / \textbf{83.9}         & 71.5 / 58.1   & 3.7 / 10.5         & 8.0 / 13.5             \\
{CCAT~\cite{ccat}}   & 93.2 / 68.9        & 74.8 / 49.8   & 2.4 / 9.9       & 7.9 / 16.1              \\
{AT + LS}   & 93.1 / 83.5        & 71.7 / 57.9 &  2.1 / 7.9     & 4.4 / 6.7               \\
{AT + multiple $\epsilon$}   & 94.3 / 84.5 & 74.6 / 59.6  &  3.4 / 10.0       & 7.0 / 12.9          \\
{AT + \al}   & \textbf{94.6} / 83.6        & \textbf{75.8} / \textbf{59.9}    & \textbf{2.0} / \textbf{6.5}      & \textbf{3.7} / \textbf{6.2}             \\
\midrule
\bottomrule 
\end{tabular*}
\end{table*}
\begin{figure*}
    \centering
    \includegraphics[width=0.49\linewidth]{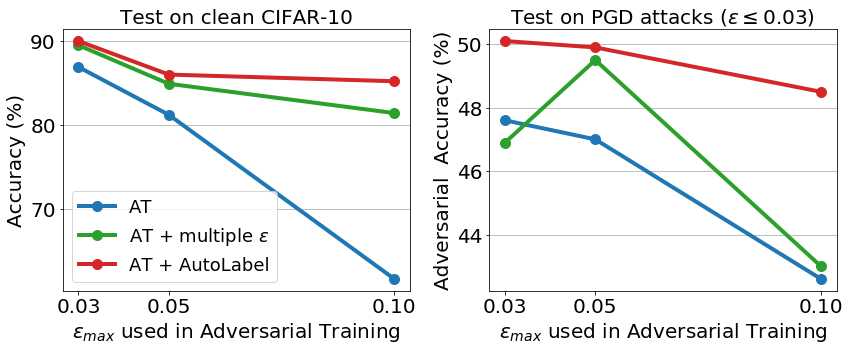} 
    \includegraphics[width=0.49\linewidth]{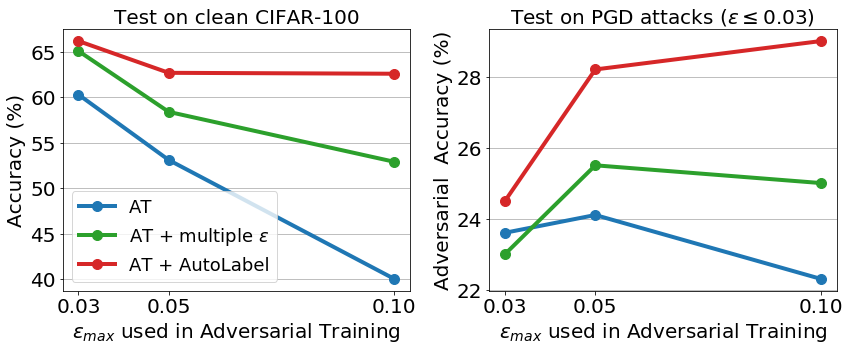} 
      
    \caption{\textbf{\al improves trade-off between clean and adversarial accuracy.} Each model is tested on the clean data (Left) and against white-box PGD attacks bounded by $\epsilon_{max} = 0.03$ (right).} 
    \label{fig:adv_al}
  
\end{figure*}

\subsubsection{\al improves calibration of adversarial training} Similar to human-designed data augmentation, we also believe adversarial training runs the risk of attacks not being truly label-preserving~\cite{qin2020deflecting} and thus should benefit from setting labels carefully. Therefore, we would like to see if \al can improve adversarial training with adversarial robustness. %

To this end, we apply \al to adversarial training and report the calibration error on the clean and corrupted datasets in Table~\ref{tab:at}. When comparing standard adversarial training (AT) with a vanilla model, we can see that adversarial training slightly improves the calibration on the corrupted dataset but at the cost of a larger calibration error on the clean data. All other adversarial training based models using one-hot labels suffer from this trade-off between clean and corrupted calibration. In contrast, \al nicely addresses this problem: it improves calibration on both clean and corrupted datasets compared to vanilla model and adversarial training. Note that a pre-defined function to smooth the labels, e.g, the power transition used in CCAT~\cite{ccat}, does not significantly help model's calibration. Although adversarial training together with \al still suffers from a small sacrifice on in-distribution accuracy, the improvements over both accuracy and calibration on the corrupted datasets are much more significant, e.g., the improvement of \al over calibration error on the corrupted data is around 60$\%$ compared to vanilla model on CIFAR-10/100-C. 

Taking all together, we can naturally arrive at the conclusion: \al can again effectively help adversarial training improve models' calibration by adaptively setting appropriate labels on the augmented (adversarial) examples.

\begin{table*}[t]
    \centering
\caption{\textbf{\al helps balance the tradeoff between clean accuracy and adversarial accuracy.}  ``Adversarial Accuracy'' denotes the model is tested on white-box PGD attacks bounded by $\epsilon_{max} = 0.03$ and generated by 50 iterations and 3 random restarts. Larger Accuracy Difference $\Delta$ is better (the arrow indicates higher is better). The best model with the highest $\Delta$ is highlighted in \textbf{bold}.}\label{tab:adv_cifar100}
\begin{tabular*}{0.96\textwidth}{c@{\extracolsep{\fill}} ccccccc}
\toprule\midrule
\multirow{3}{*}{\textbf{Method}}  & \multirow{3}{*}{\begin{tabular}[c]{@{}c@{}}$\mathbf{\epsilon_{max}}$ in\\ \textbf{ Adversarial Training}\end{tabular}} & \multicolumn{3}{c}{\textbf{CIFAR-10}} & \multicolumn{3}{c}{\textbf{CIFAR-100}}\\
\cmidrule(r){3-5}  \cmidrule(r){6-8} 
        && {\begin{tabular}[c]{@{}c@{}}Clean\\ Accuracy\end{tabular}} &  {\begin{tabular}[c]{@{}c@{}} Adversarial\\ Accuracy\end{tabular}} & $\Delta$ ($\uparrow$) & {\begin{tabular}[c]{@{}c@{}}Clean\\ Accuracy\end{tabular}} &  {\begin{tabular}[c]{@{}c@{}} Adversarial\\ Accuracy\end{tabular}} & $\Delta$ ($\uparrow$) \\ \midrule
        Vanilla & - & 95.6 & 0 & - & 79.5 & 0 & - \\ \midrule
         \multirow{3}{*}{AT} & $0.03$ & 86.9 & 47.6 &  + 38.9 & 60.3 & 23.6 &  + 4.4\\ 
         &  $0.05$ & 81.2 & 47.0 &  + 32.6& 53.1 & 24.1 &  - 2.3\\ 
         & $0.1$ & 61.6 & 42.6 & + 8.6  & 40.0 & 22.3 & -17.2\\ \midrule
          \multirow{3}{*}{CCAT} & $0.03$ & 92.9 & 0 &- 2.7& 73.3 & 0 &- 6.2 \\ 
         &  $0.05$ & 92.4 & 1.9 & - 1.3& 72.3 & 0.8 & - 6.4\\ 
         & $0.1$ & 92.1 & 4 & + 0.5& 71.2 & 2.8 & - 5.5\\ \midrule
          \multirow{3}{*}{AT + LS} & $0.03$ & 86.7 & 52.3 & + 43.4  & 61.2 & 27.4 & + 9.1\\ 
         &  $0.05$ & 81.9 & 46.6 & + 32.9& 53.7 & 26.7 & + 0.9\\ 
         & $0.1$ & 61.0 & 36.2 & + 1.6 & 39.6 & 23.8 & -16.1\\ \midrule
          \multirow{3}{*}{{\begin{tabular}[c]{@{}c@{}}AT + \\ multiple $\epsilon$\end{tabular}}} & $ 0.03$ & 89.5 & 46.9 & + 40.8& 65.1 & 23.0 & + 8.6\\ 
         &  $0.05$ & 84.9 & 49.5 & + 38.8& 58.4 & 25.5 & + 4.4 \\ 
         & $0.1$ & 81.4 & 43.0 & + 28.8 & 52.9 & 25.0 & - 1.6 \\ \midrule
          \multirow{3}{*}{{\begin{tabular}[c]{@{}c@{}}AT + \\ \al\end{tabular}}} & $0.03$ &90.0 &50.1 & \textbf{+ 45.5} & 66.2 &24.5 & \textbf{+ 11.2}\\ 
         &  $0.05$ &86.0& 49.9 & \textbf{+ 40.3}&62.7 & 28.2 & \textbf{+ 11.4}\\
         & $ 0.1$ & 85.2 & 48.5 & \textbf{+ 38.1} & 62.6 & 29.0 & \textbf{+ 12.1}\\ \midrule
         \bottomrule
    \end{tabular*}
   
    \label{tab:adv_cifar10}
\end{table*}
\vspace{3mm}
\section{AutoLabel improves adversarial robustness}

In this section, we further investigate if \al can also help bridge trade-off between clean accuracy \& adversarial robustness, especially when highly distorted adversarial examples are incorporated into training.
\subsection{Baselines} We use the same baselines introduced in Section~\ref{sec:baseline_at}. The ``Vanilla'' model is trained without any adversarial examples. For other adversarial training based methods, we construct $\ell_\infty$ norm based PGD attacks with 10 iterations and the step size is set to be $\epsilon/4$. For each method, we train three models with the $\ell_\infty$ norm of adversarial perturbation $\epsilon_{max}$ chosen from $\{0.03, 0.05, 0.1\}$, where the image scale is [0, 1]. 

\subsection{Evaluation Metrics} We report both clean accuracy and adversarial accuracy of each test model. The adversarial accuracy is computed over white-box PGD attacks bounded by $\epsilon_{max} = 0.03$ with image scale belongs to [0, 1]. These test PGD attacks are generated by 50 iterations and 3 random restarts. 

In addition, we also propose a new evaluation metric called ``Accuracy Difference $\Delta$'' to better measure the effectiveness of any given method in terms of clean accuracy as well as adversarial robustness. ``Accuracy Difference $\Delta$'' is to compute the clean and adversarial accuracy difference between any given method as well as a baseline model. In our case, the ``Vanilla'' model trained with one-hot labels is used as the baseline. Denote any given method as $\Omega$ and the ``Vanilla'' model as $\Gamma$, the ``Accuracy Difference $\Delta$'' between $\Omega$ and $\Gamma$ can be computed as
\begin{equation}
\begin{split}
   \Delta(\Omega, \Gamma) & =  \{\textrm{Accuracy}(\Omega) + \textrm{Adversarial. Accuracy}(\Omega)\}\\ 
   & ~~~ -  \{\textrm{Accuracy}(\Gamma)  + \textrm{Adversarial. Accuracy}(\Gamma)\}
\end{split}
\end{equation}
 where the accuracy and adversarial accuracy are computed on the clean test dataset as well as white-box PGD attacks respectively. While computing $\Delta$, clean accuracy and adversarial accuracy are assigned with equal importance. If $\Delta(\Omega, \Gamma)$ is larger than 0, this indicating that this method $\Omega$ is overall better than the vanilla model $\Gamma$ in terms of clean accuracy as well as adversarial robustness. 

\subsection{Network Architectures and Hyperparameters}
We use the same Wide ResNet-28-10~\cite{Zagoruyko2016WideRN} for both CIFAR-10 and CIFAR-100 datasets introduced in Section~\ref{sec:arch}. 

For adversarial training together with label smoothing (AT+ LS), we sweep the smoothing parameter $\rho$ from $\{0.1, 0.2, 0.3\}$ and choose the best parameter for each $\epsilon_{max}$ on CIFAR-10 and CIFAR-100. Similarly, for adversarial training with \al, we set the number of distance buckets to be $N=10$ and sweep the hyperparameter $\alpha$ in Eqn.~(\ref{eqn:update}) within a range [0.005, 1] and choose the best based on the performance on the holdout validation set.

\subsection{Improvements over Adversarial Robustness} To validate if \al can help balance clean accuracy and adversarial accuracy when highly distorted adversarial examples are incorporated into training, we compare \al with standard adversarial training (AT)~\cite{madry2017towards} as well as adversarial training with multiple $\epsilon$ (AT + multiple $\epsilon$), which are trained with one-hot labels. 
We see a clear picture in Figure~\ref{fig:adv_al}: as the $\epsilon_{max}$ used in adversarial training increases, that is we incorporate more corrupted adversarial examples into training, there is a significant clean accuracy drop for the models trained with one-hot labels. In contrast, \al effectively helps the model maintain a relatively high clean accuracy as well as adversarial accuracy even when $\epsilon_{max} = 0.1$. 

This picture is even more clear when we look at the Accuracy Difference $\Delta$ in Table \ref{tab:adv_cifar100}: AT + \al keeps $\Delta$ relatively stable and always larger than 0 when we increase $\epsilon_{max}$ used in adversarial training, e.g., $\epsilon_{max} = 0.1$, whereas other models suffer from a significant performance drop. This further validates that \al is especially useful when highly corrupted adversarial examples are involved into training. Note this is perfectly aligned with the observation in standard data augmentation techniques shown in Figure~\ref{fig:acc_ece}: assigning one-hot labels to highly-distorted augmented data can hurt accuracy and \al can better balance the accuracy trade-off by assigning more appropriate labels for augmented data.

In addition, since confidence-calibrated adversarial training (CCAT)~\cite{ccat} is originally proposed to detect adversarial examples with a confidence-thresholding, it hardly helps adversarial accuracy compared to vanilla model, as shown in Table~\ref{tab:adv_cifar100}. This is mainly because CCAT uses the uniform distribution for adversarial examples during training, resulting in a loss of class information.

\section{Conclusion}
In this paper, we show that simply re-using one-hot labels for augmented data, as commonly used in existing data augmentation works, runs the risk of adding noise and degrading accuracy and calibration. To mitigate this, we propose \al to automatically learn the confidence in labels for augmented data based on the transformation distance between the augmented data and the clean data.
We demonstrate the effectiveness of \al by applying it to RandAug, AugMix and adversarial training. We see that \al greatly improves the models' calibration, especially on corrupted data, and also helps adversarial training with a better trade-off between clean accuracy and adversarial robustness.  More generally, we believe that more nuanced approaches to setting labels for augmented data, beyond assuming label-preserving transformations, will lead to more effective data augmentation techniques.

\bibliographystyle{IEEEtran}  
\bibliography{main}

\appendix

\subsection{AutoLabel for mixup}
Other than data augmentation techniques that using one-hot labels, we also apply \al to mixup~\cite{Zhang2018mixupBE}, which mixes the input data as well as their correspondingly labels.
Recently, mixup has also been shown to be able to help with calibration in~\cite{Thulasidasan2019OnMT}. Specifically, the input data as well as their labels are augmented by 
\begin{equation}\label{eqn:mixup}
\begin{split}
  \aug_{mixup}(x_i, x_j) = \gamma x_{i} + (1 - \gamma) x_{j} \\
  \aug_{mixup}(y_i, y_j) = \gamma \hat{y}_{i} + (1 - \gamma) \hat{y}_{j}
\end{split}
\end{equation}
where $x_i$ and $x_j$ are two randomly sampled input data and $\hat{y}_i$ and $\hat{y}_j$ are their associated one-hot labels. The model is trained with the standard cross-entropy loss $\mathcal{L}(f(x_{mixup}), {y}_{mixup})$ and the mixing parameter $\gamma \in [0, 1]$ that determines the mixing ratio is randomly sampled from a Beta distribution Beta($\beta, \beta$) at each training iteration. Rather than using the same mixing parameter $\gamma$ to combine the labels, we show how to apply \al to automatically learn its labels $ \aug_{mixup}(y_i, y_j) $ based on the validation calibration.

\paragraph{Transformation Distance} 
The transformation distance in mixup is determined by the mixing parameter $\gamma$ in Eqn (\ref{eqn:mixup}). When $\gamma \rightarrow 0.5$, combining two images equally, the augmented image $\aug_{mixup}(x_i, x_j)$ is the most far away from the clean distribution. On the other hand, when $\gamma \rightarrow 0$, the augmented image is close to the original image.
Hence, the distance bucket $S_n$ for each augmented example $\aug_{mixup}(x_i, x_j)$ can be defined as:
    $S_n = S_{\lceil 2N \cdot (\min(\gamma, 1 - \gamma)) \rceil}.$
\paragraph{Update Labels} To learn the labels for the augmented training image within the distance bucket $S_n$, \al constructs an augmented validation set $\mathcal{Q}(S_n)$ by randomly mixing two images from validation data with a mixing parameter $\gamma'$ that is sampled from a uniform distribution: $\gamma' \sim \mathcal{U}\left(\frac{n}{2N}, \frac{n + 1}{2N}\right)$ and $\gamma' \in [0, 0.5]$. Unlike AugMix, there are two classes $y_i$ and $y_j$ existing in the augmented image $\aug_{mixup}(x_i, x_j)$. Due to $\gamma' \in [0, 0.5]$, the class in the image $x_j$ plays a dominant role in determining the main class in $\aug_{mixup}(x_i, x_j)$. Therefore, we follow Eqn (1) in the main text to update the label $\Tilde{y}_{k=y_j}$ for the class $k=y_j$. Unlike Eqn~(2) in the main text that uniformly distributes the probability $1 - \Tilde{y}_{k=y_j}$ to all other classes, we update the label for the class $k=y_i$ as $\Tilde{y}_{k=y_i}  = \min( 1 - \Tilde{y}_{k=y_j}, \frac{\gamma'}{1 -\gamma'}\Tilde{y}_{k=y_j})$ and then distribute the probability $1- \Tilde{y}_{k=y_i} -\Tilde{y}_{k=y_j}$ to all other $K-2$ classes.
Finally, the model is trained by minimizing the cross-entropy loss with the new labels $\tilde{y}$ as the target.
\subsection{Experiments on mixup}
\subsubsection{Setup} When applying \al to mixup, we set the number of distance buckets to be $N = 5$. The hyperparameter $\alpha$ in Eqn (1) is sweep in a set and we choose the best $\alpha = 0.005$ for CIFAR10 and $\alpha = 0.008$ for CIFAR100.

\subsubsection{Results analysis} To test how well \al improves mixup \cite{Zhang2018mixupBE} is more nuanced because mixup's baseline effectiveness is  sensitive to its hyperparameters.
In particular, the mixing parameter $\gamma$ in Eqn~(\ref{eqn:mixup}) is sampled from a beta distribution Beta$(\beta, \beta)$. When $\beta \rightarrow 0$, most sampled $\gamma$ are close to 0 or 1; when $\beta = 1$, $\gamma$ is randomly sampled from a uniform distribution. We observe that mixup suffers from a trade-off between accuracy and calibration on the clean data on CIFAR10 and CIFAR100, shown in Table~\ref{tab: mixup_clean}. When the hyperparameter $\beta$ is large, e.g., $\beta=1$, the model is trained on more diverse augmented data compared to a smaller $\beta$, e.g., $\beta=0.2$. This results in a higher accuracy but leads the model to be too under-confident and a much larger calibration error on the clean data. This trade-off between clean accuracy and calibration of mixup is also observed in other datasets and networks in Figure 2(j) in \cite{Thulasidasan2019OnMT}. After applying \al to mixup to automatically adjust the labels for augmented images, we find that the trade-off between accuracy and calibration is well addressed: high accuracy and low calibration error are achieved on the clean data, shown in Table~\ref{tab: mixup_clean}. However, this trade-off between accuracy and calibration does not exist in the large scale datasets, e.g., ImageNet, where $\beta=0.2$ consistently has the best accuracy and calibration and we did not observe a significant improvement when applying \al to mixup on ImageNet.

In addition, we also find that \al can improve accuracy and calibration of mixup when applying \al to mixup with $\beta=1$ on the corrupted datasets, as shown Table~\ref{tab: mixup_corrupt}. As we can see that the mean accuracy on CIFAR100-C over mixup with from 55.6$\%$ to 56.9$\%$ and reduce the mean calibration error from $11.2\%$ to $10.0\%$.

\begin{table}[bt!]
\caption{Effects of \al for mixup on CIFAR10 and CIFAR100. All numbers reported are averaged over 4 independent runs and in $\%$. Best highlighted in \textbf{bold}.}\label{tab: mixup_clean}
\centering
\small
\begin{tabular}{ccccc}
\toprule
 \multirow{2}{*}{\textbf{Method}}     & \multicolumn{2}{c}{\textbf{CIFAR10}}              & \multicolumn{2}{c}{\textbf{CIFAR100}} \\ \cmidrule(r){2-3} \cmidrule(r){4-5}
                                                                      & {Accuracy} & ECE       & Accuracy        & ECE        \\ \midrule
\textbf{Vanilla}                                                                & {95.6}     & 2.6       & 79.5            & 6.1        \\ \midrule
\textbf{mixup} ($\beta$ = 0.2)                                                     &{96.2}     & 0.8       & 80.8            & 1.8        \\
\textbf{mixup} ($\beta$ = 1)                                                       & {96.5}     & 5.3       & 80.9            & 5.5        \\\midrule
\textbf{+ \al} \\($\beta$ = 1)& \textbf{96.7}    & \textbf{0.6}       & \textbf{81.1}            & \textbf{1.2}       \\ \bottomrule
\end{tabular}
\end{table}

\begin{table}[bt!]
\caption{Effects of \al for mixup on the corrupted datasets: CIFAR10-C and CIFAR100-C. All numbers reported are averaged over 4 independent runs and in $\%$. Best highlighted in \textbf{bold}.}\label{tab: mixup_corrupt}
\centering
\small
\begin{tabular}{ccccc}
\toprule
 \multirow{2}{*}{\textbf{Method}}     & \multicolumn{2}{c}{\textbf{CIFAR10-C}}              & \multicolumn{2}{c}{\textbf{CIFAR100-C}} \\ \cmidrule(r){2-3} \cmidrule(r){4-5}
                                                                      & {Accuracy} & ECE       & Accuracy        & ECE        \\ \midrule

\textbf{mixup}      & {81.1}     & 8.8      & 55.6          &11.2  \\\midrule
\textbf{+ \al} & \textbf{81.3}    & \textbf{8.5}       & \textbf{56.9}            & \textbf{10.0}       \\ \bottomrule
\end{tabular}
\vspace{-2mm}
\end{table}

\subsection{Updates based on previous reviews}
\subsubsection{Previous Meta Reviews} The paper introduces a simple and interesting method that adaptively smoothes the labels of augmented data based on a distance to the “clean” training data. The reviewers have raised concerns about limited novelty, minor improvement over baselines, and insufficient experiments. The author’s response was not sufficient to eliminate these concerns. The AC agrees with the reviewers that the paper does not pass the acceptance bar of ICLR.

\subsubsection{Updates} There is a fundamental change of the new submission compared to the previous version in terms of methodology as well as empirical experiments. We apply \al to more types of data augmentation to discover its effectiveness for models' calibration and adversarial robustness, especially when highly corrupted augmented data are incorporated into training.
\end{document}